\documentclass[lettersize, journal]{IEEEtran}
\usepackage{amsmath, amsfonts}
\usepackage{algorithmic}
\usepackage{orcidlink}
\usepackage{algorithm}
\usepackage{CJKutf8}
\usepackage{array}
\usepackage[caption=false, font=normalsize, labelfont=sf, textfont=sf]{subfig}
\usepackage{textcomp}
\usepackage{stfloats}
\usepackage{url}
\usepackage{verbatim}
\usepackage{amssymb}
\usepackage{graphicx}
\usepackage{cite}
\usepackage{xcolor}
\begin{document}
\begin{CJK*}{UTF8}{gbsn}

\title{Captioning Daily Activity Images in Early Childhood Education: Benchmark and Algorithm}

\author{
Sixing Li\,\orcidlink{0009-0005-1033-2329}%
\thanks{Sixing Li and Zhibin Gu contributed equally to this work.}%
,
Zhibin Gu\,\orcidlink{0000-0002-1085-9084}\footnotemark[1],
Ziqi Zhang\,\orcidlink{0000-0002-5937-183X},
Weiguo Pan\,\orcidlink{0000-0002-2293-1004},
Bing Li\,\orcidlink{0000-0002-5888-6735},
Ying Wang,
and Hongzhe Liu\,\orcidlink{0000-0003-2314-5272}%

\thanks{
Sixing Li, Weiguo Pan, and Hongzhe Liu are with the School of Robotics, 
Beijing Union University, Beijing 100101, China 
(e-mail: lisixing@buu.edu.cn; ldtweiguo@buu.edu.cn; liuhongzhe@buu.edu.cn).
}

\thanks{
Zhibin Gu is with the College of Computer and Cyber Security, 
Hebei Normal University, Shijiazhuang 050010, China 
(e-mail: guzhibin@hebtu.edu.cn).
}

\thanks{
Ziqi Zhang is with the Institute of Automation, Chinese Academy of Sciences (CASIA), Beijing 100190, China (e-mail: zhangziqi2017@ia.ac.cn)
}

\thanks{
Bing Li is with the National Laboratory of Pattern Recognition (NLPR),
Institute of Automation, Chinese Academy of Sciences (CASIA), Beijing
100190, China, also with the School of Artificial Intelligence, University of
Chinese Academy of Sciences (UCAS), Beijing 100190, China, and also with
PeopleAI, Inc., Beijing 100190, China (e-mail: bli@nlpr.ia.ac.cn).
}
\thanks{
Wangying is with People Youhe Education Technology Co., Ltd., Shandong 250033, China 
(e-mail: 15210019181@163.com).
}
\thanks{
This work was supported in part by the National Natural Science Foundation of China under Grant 62506116, and in part by the National Natural Science Foundation of China under Grant U24A20331.
}

\thanks{
Corresponding author: Hongzhe Liu (e-mail: liuhongzhe@buu.edu.cn).
}
}



\maketitle

\begin{abstract}
Image captioning for Early Childhood Education (ECE) is essential for automated activity understanding and educational assessment. However, existing methods face two key challenges. First, the lack of large-scale, domain-specific datasets limits the model’s ability to capture fine-grained semantic concepts unique to ECE scenarios, resulting in generic and imprecise descriptions\cite{rcmf}. Second, conventional training paradigms exhibit limitations in enhancing professional object description capability, as supervised learning tends to favor high-frequency expressions, while reinforcement learning may suffer from unstable optimization on difficult samples.

To address these limitations, we introduce ECAC, a large-scale benchmark for ECE daily activity image captioning, comprising 256,121 real-world images annotated with expert-level captions and fine-grained labels. ECAC is further equipped with a domain-oriented evaluation protocol, the Teaching Toy Recognition Score (TTS), to explicitly measure professional object naming accuracy. Furthermore, we propose RSRS (Reward-Conditional Switch of Reinforcement Learning and Supervised Fine-Tuning), a hybrid training framework that dynamically alternates between RL and supervised optimization. By rerouting hard samples with zero rewards to supervised fine-tuning, RSRS effectively mitigates advantage collapse and enables stable optimization for fine-grained recognition. Leveraging ECAC and RSRS, we develop KinderMM-Cap-3B, a domain-adapted multimodal large language model. Extensive experiments demonstrate that our model achieves a TTS of 51.06, substantially outperforming state-of-the-art baselines while maintaining superior caption quality, highlighting its potential for specialized educational applications. The code and data are available at https://github.com/SixingLI030/KinderMM-Cap.
\end{abstract}

\begin{IEEEkeywords}
Early childhood education, Image captioning,Reinforcement learning, Supervised fine-tuning,Multimodal language models
\end{IEEEkeywords}

\section{Introduction}
\IEEEPARstart{M}{ultimodal} large language models (MLLMs)\cite{fla},\cite{mini} integrate large language models (LLMs)\cite{llm},\cite{tlm} with the ability to process multiple modalities such as text, images, audio, and video. Their core lies in aligning and representing multimodal information, typically using language as a bridge for cross-modal understanding and generation. Compared to traditional text-only models, MLLMs better mimic human cognition and show strong applicability across domains, including medical imaging and diagnosis\cite{hfb},\cite{pml}, financial analysis\cite{fin},\cite{mme},\cite{cfb}, fashion product description\cite{dfp}, and agricultural intelligence\cite{agr}.

Multimodal data are widely prevalent in educational scenarios. For instance, in multimedia classrooms, video\cite{isy}, dialogue\cite{etp}, and speech\cite{etpo} data during teaching processes can be collected to depict teaching activities from multiple perspectives. Owing to the capabilities of MLLMs in cross-modal alignment and joint semantic modeling, such models sparked widespread interest in exploring ways to leverage them for effectively mining multimodal information in the education sector to improve teaching quality. For example,  Lee et al. \cite{ner} and Lee et al. \cite{gpd} analyzed teaching processes and student behaviors based on classroom records, while a study \cite{doc} further explored the potential of promoting personalized learning through customized content generation and interactive multimodal dialogue. Among various educational scenarios, Early Childhood Education (ECE) focuses on preschool children whose cognitive, linguistic, and expressive abilities are still developing, making their behavioral intentions and learning needs difficult to articulate explicitly and often latent within complex multimodal interactions. This characteristic further highlights the necessity of leveraging MLLMs for in-depth understanding and modeling of multimodal data in ECE settings. Recent studies have begun to explore this direction by employing MLLMs to analyze joint attention in parent–child interactions \cite{tml} and by integrating MLLMs into artificial intelligence peer systems to enable context-aware and emotion-responsive interactions \cite{dino}.

However, existing studies predominantly focus on conversational interaction and companion oriented functions, while research on image description tailored to daily activities in preschool educational settings remains relatively limited\cite{aat},\cite{mde}. Image description is a cross-modal task that converts visual content in images into semantically consistent natural language, with its preschool education-oriented variant focusing on depicting children’s daily activities to reflect their behavioral and developmental traits. Advancing image-based daily activity understanding 
through image captioning therefore provides a principled pathway toward scalable and objective 
analysis in this domain.

Despite its potential, image captioning in ECE scenarios faces two fundamental challenges. First, the scarcity of publicly available image captioning datasets tailored to ECE scenarios imposes a fundamental constraint on model training. Without such domain-specific datasets, models are unable to internalize the unique semantic information of ECE scenarios, including concepts related to specialized educational toys and functional activity areas, and thus fail to achieve the accurate visual-language alignment essential for effective image captioning. Second, limited by conventional training paradigms, multimodal models often generate vague or non-professional descriptions when depicting domain-specific objects after supervised fine-tuning, failing to meet the accuracy requirements of early childhood education. This issue mainly arises from two factors. On the one hand, the conflict between general visual–language knowledge learned during large-scale pre-training and domain-specific semantics introduced during fine-tuning leads models to prefer generic or high-frequency object names. Furthermore, educational toys in ECE environments often exhibit high visual ambiguity due to small object scale, partial occlusion, and inter-class similarity. These factors increase recognition difficulty and bias models toward generic descriptions rather than precise terminology. To mitigate this, multi-scale spatial feature extraction methods like Spatial Pyramid Transformers\cite{spt} have been explored.

\begin{figure*}
    \centering
    \includegraphics[width=\textwidth]{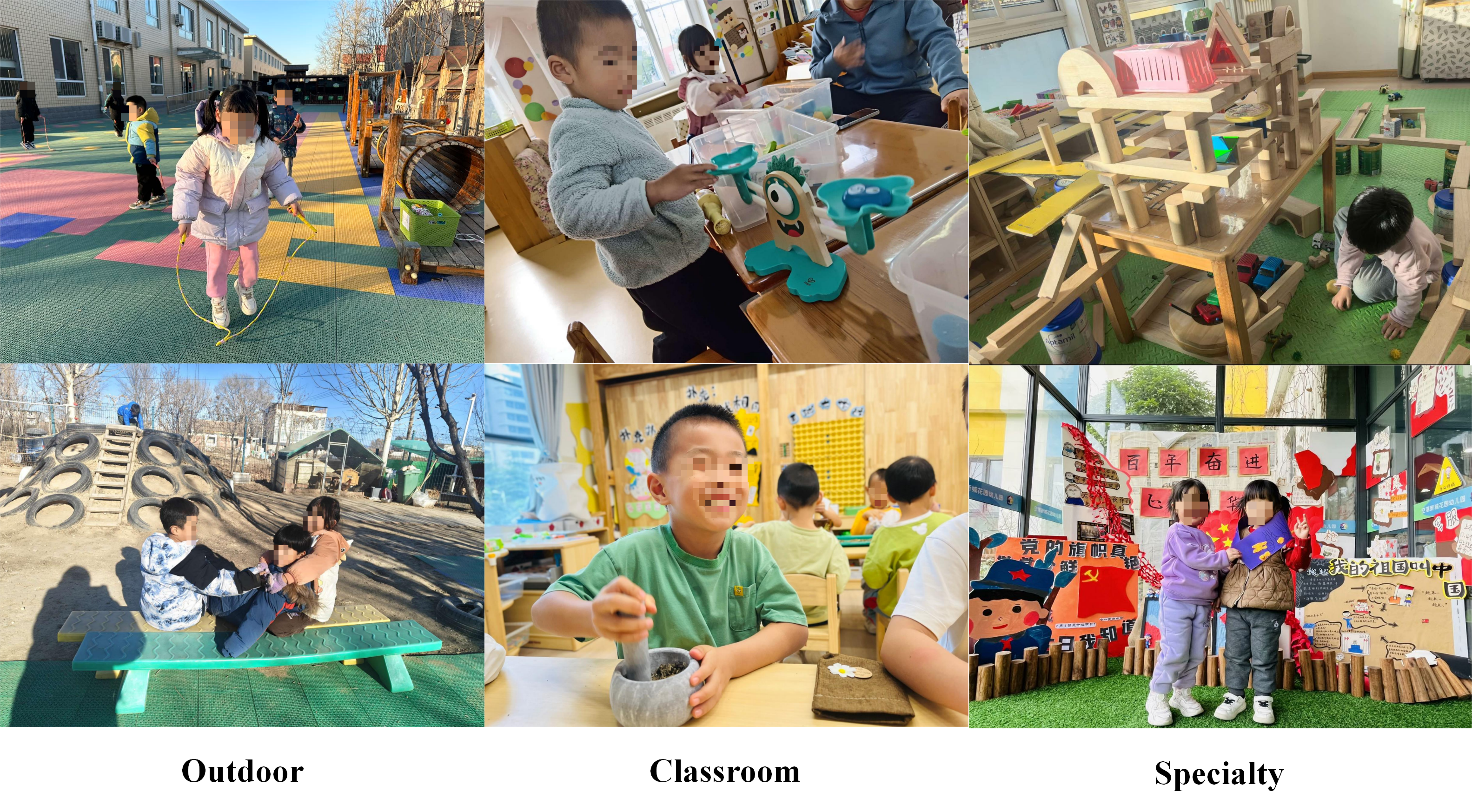} 
    \caption{Daily Activity Images Across Three Scenario Categories}
    \label{fig:image_cases}
\end{figure*}

\IEEEpubidadjcol
To address the above challenges, we tackle the problem from both the data and training paradigm perspectives. On the data side, we construct ECAC, a large-scale benchmark dataset for daily activity image captioning in early childhood education. ECAC consists of 256,121 real-world images collected from authentic kindergarten environments and is annotated with expert-level captions that emphasize pedagogically relevant content. In addition to free-form descriptions, ECAC provides fine-grained annotations of teaching toys and activity areas, enabling models to learn domain-specific terminology and visual–semantic associations that are absent from existing generic captioning datasets.On the training side, to explicitly address the vague description issue caused by knowledge conflicts and visual ambiguity, we propose RSRS (Reward-Conditional Switch of Reinforcement Learning and Supervised Fine-Tuning), a hybrid optimization framework designed for fine-grained, domain-specific captioning. Unlike conventional reinforcement learning strategies that suffer from advantage collapse when all candidate descriptions receive identical or zero rewards, RSRS dynamically switches the optimization path based on reward distributions. Hard samples that fail to trigger discriminative rewards are routed into supervised fine-tuning, allowing the model to directly learn from expert annotations, while other samples continue to benefit from reinforcement-driven exploration. This design enables RSRS to stabilize training and progressively improve the professional accuracy of object naming without sacrificing overall linguistic quality.The main contributions of this work are as follows:
\begin{itemize}
    \item We introduce a benchmark dataset of daily activities in ECE settings, along with tailored evaluation metrics and methodologies for assessing the quality of image captions of daily activities in ECE.
    \item We designed a novel training framework that dynamically integrates reinforcement learning and supervised fine-tuning under reward control, effectively enhancing fine-grained expression in image captions.
    \item We trained a multimodal large language model to generate high-quality captions for images in ECE settings, with a focus on accurately identifying classroom areas and describing educational toys.
\end{itemize}

\begin{table*}[t] 
    \centering
    \caption{Different descriptions of the same educational toy}
    \renewcommand{\arraystretch}{1.6}
    \label{tab:compare_case}
    \begin{tabular}{l>{\raggedright\arraybackslash}p{0.4\textwidth}>{\raggedright\arraybackslash}p{0.4\textwidth}}
    \hline
    &\vspace{0.1mm}\includegraphics[width=0.35\textwidth]{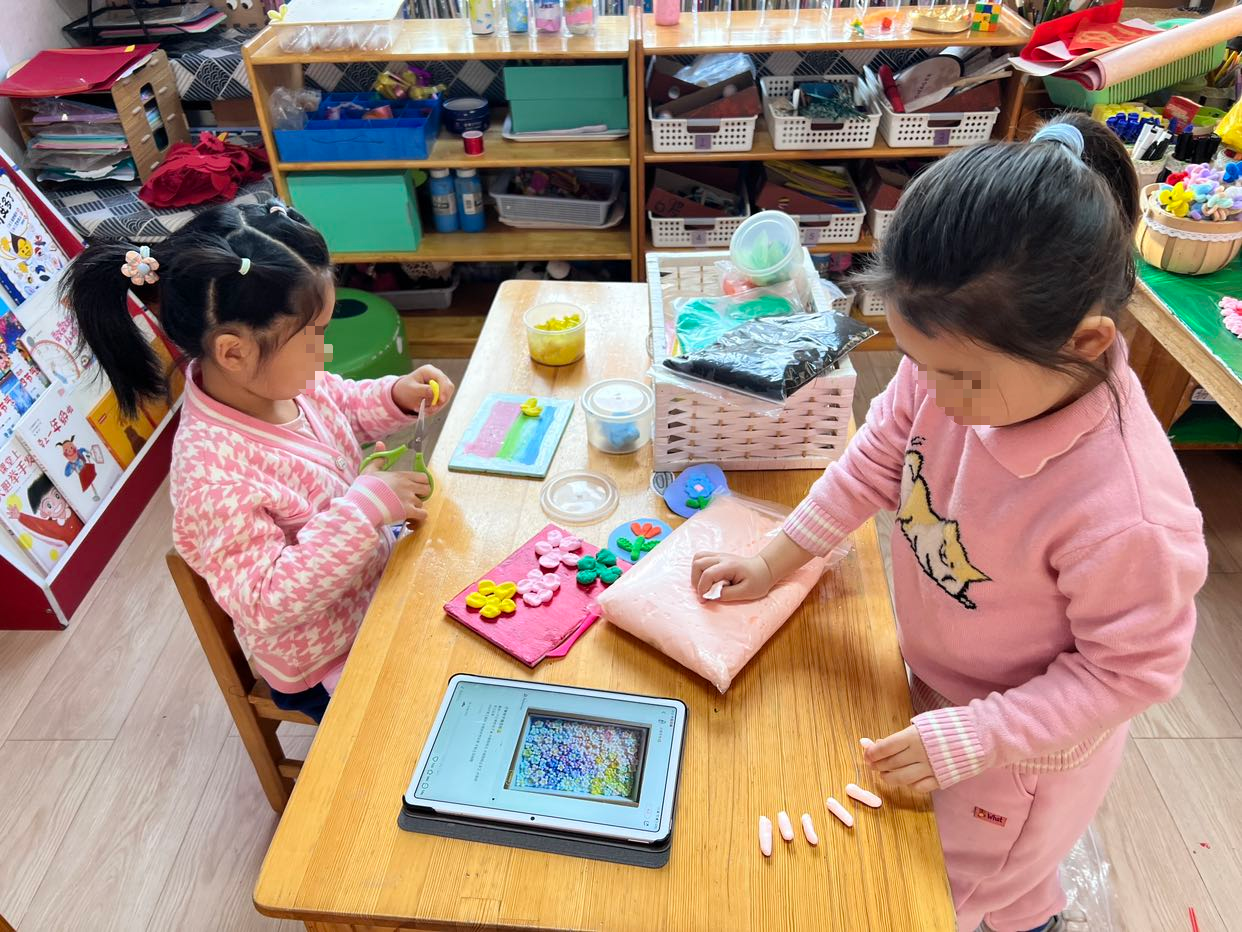}&\vspace{0.1mm}
    \includegraphics[width=0.35\textwidth]{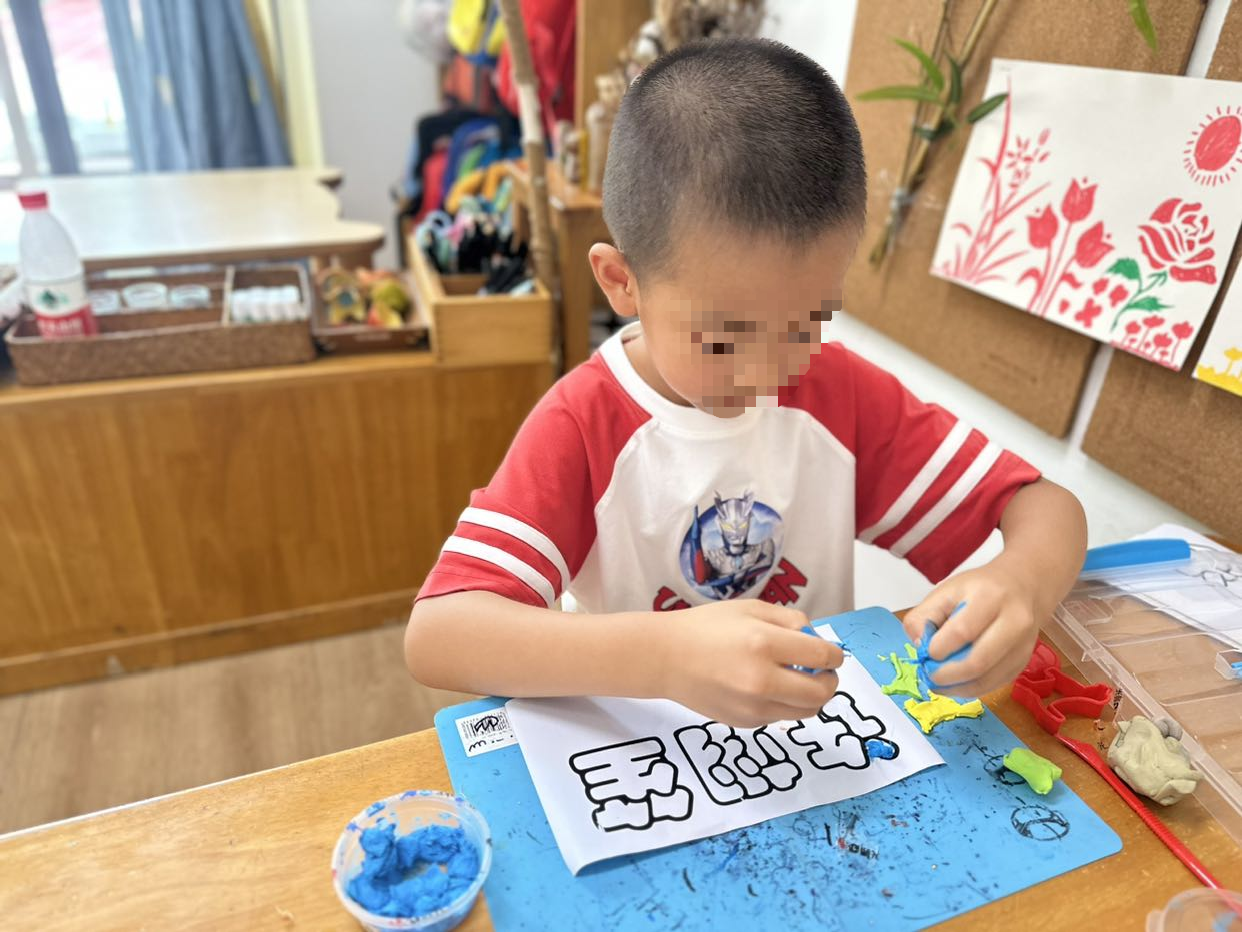} \\
    \hline
    Prompt & 
    根据图片定位到幼儿园的具体场景，根据这张图片内容生成一段流畅且连贯的对主要内容的描述。我希望句子是简洁但包含关键细节，例如发生的具体场景，主要人物的动作、表情以及精确识别与主要人物有关的物品。不要描述与主要人物无关的背景，不要描述与当前主要活动无关的物体，不要描述没有露脸或与画面主要内容无关的人物，不要描述穿着的日常衣服（特殊装扮需要描述），不要描述不必要的表情。(Locate the specific kindergarten scene based on the image, and generate a smooth and coherent description of its main content according to the image. I want the sentences to be concise but include key details, such as the exact setting, the main characters’ actions and expressions, and accurate identification of objects related to the main characters. Do not describe background elements unrelated to the main characters, objects irrelevant to the main activity, people whose faces are not visible or who are unrelated to the main content, ordinary clothing (describe only special costumes), or unnecessary expressions.)
    & 
    根据图片定位到幼儿园的具体场景，根据这张图片内容生成一段流畅且连贯的对主要内容的描述。我希望句子是简洁但包含关键细节，例如发生的具体场景，主要人物的动作、表情以及精确识别与主要人物有关的物品。不要描述与主要人物无关的背景，不要描述与当前主要活动无关的物体，不要描述没有露脸或与画面主要内容无关的人物，不要描述穿着的日常衣服（特殊装扮需要描述），不要描述不必要的表情。(Locate the specific kindergarten scene based on the image, and generate a smooth and coherent description of its main content according to the image. I want the sentences to be concise but include key details, such as the exact setting, the main characters’ actions and expressions, and accurate identification of objects related to the main characters. Do not describe background elements unrelated to the main characters, objects irrelevant to the main activity, people whose faces are not visible or who are unrelated to the main content, ordinary clothing (describe only special costumes), or unnecessary expressions.) \\
    \hline
    Generation & 
    班级美工区中，两个小女孩正在桌子上专注地进行手工制作。左边的小女孩穿着粉色带花图案的上衣，她正用手摆弄着\textcolor{red}{桌上的材料}；右边的小女孩穿着粉色上衣，上面有卡通图案，她也在认真地操作着\textcolor{red}{手中的物品}。桌上摆放着各种\textcolor{red}{手工材料}、一台显示着画面的平板电脑以及一些已经制作好的\textcolor{red}{小物件}，她们似乎很享受这充满创意的手工制作过程。(In the classroom art corner, two little girls are  intently engaged in hand-icraft work at the table. The girl on the left is wearing a pink top with floral patterns, and she is fiddling with \textcolor{red}{the materials} on the table. The girl on the right is wearing a pink top with cartoon patterns, and she is also carefully working with \textcolor{red}{the item} in her hands. On the table are various \textcolor{red}{handicraft materials}, a tablet computer displaying an image, as well as \textcolor{red}{some small items} that have already been made. They seem to be thoroughly enjoying this creative handicraft-making process.)
    
    &
    在班级美工区，一个小男孩正坐在铺有花卉图案桌布的桌子旁，手里拿着一块\textcolor{green}{绿色的彩泥}，似乎正在专注地进行手工创作，他神情认真，沉浸在自己的手工世界里，周围还有其他小朋友也在进行着类似的活动。(In the classroom art corner, a little boy is sitting at a table covered with a floral-patterned tablecloth, holding a piece of \textcolor{green}{green clay} in his hand. He seems to be deeply focused on his handicraft creation, with a serious expression, fully immersed in his own world of craft-making. Around him, other children are also engaged in similar activities.) \\
    \hline
    \end{tabular}
\end{table*}
  
\section{RELATED WORKS}
\subsection{Multimodal Large Language Model}
In recent years, following the remarkable achievements of large language models (LLMs), researchers 
increasingly shifted their focus toward the study and development of MLLMs. The 
objective is to enable these models to understand and generate information across multiple 
modalities simultaneously. A multitude of groundbreaking MLLMs have emerged, such as Blip\cite{blip} and 
Qwen2.5-VL\cite{qw25}, which achieved exceptional performance across a range of multimodal 
tasks\cite{VQA},\cite{okvqa},\cite{docvqa},\cite{ocrb},\cite{msr}. 

Unlike tasks based on common images\cite{coco},\cite{oven}, some researchers have begun to explore the potential of
MLLMs in vertical domains, addressing tasks that rely on domain-specific images, such as medical imaging 
analysis\cite{med},\cite{amo}, remote sensing image question answering\cite{vrs}, and financial image 
analysis\cite{vis}. In this work, we explore the adaptation of MLLMs to the ECE and 
evaluate the effectiveness of different training approaches. The aim is to provide insights for applying MLLMs in other vertical domains.

\subsection{Applications of Computer Vision in ECE}
Computer vision has played a significant role in ECE, enabling applications such as 
emotion recognition, face tracking, and behavioral analysis. Existing studies primarily focus on leveraging facial 
expression recognition to assess children’s emotional states in real time\cite{acv}, deploying face-tracking 
systems for classroom monitoring and personalized instruction\cite{irf}, and utilizing transfer learning or deep 
neural networks to analyze fine-grained facial dynamics, such as smiles and attention patterns\cite{cul},\cite{dso}. 
Furthermore, vision-based technologies are frequently integrated into social robots  to improve interaction quality 
and facilitate therapeutic interventions for children with autism spectrum disorder (ASD) or speech 
impairments\cite{tka}.

While these efforts demonstrated the effectiveness of computer vision in ECE, most prior work concentrates on 
recognition or classification tasks rather than generating descriptive and pedagogically relevant textual content. 
To address this limitation, our work explores the adaptation of MLLMs to ECE-specific tasks, focusing on the 
automatic generation of natural language captions for daily activity images in ECE settings. By leveraging the 
cross-modal reasoning capabilities of MLLMs, our approach aims to produce contextually appropriate and 
educationally meaningful captions This not only supports classroom documentation and interactive learning but 
also enhances communication between educators and families. Furthermore, by investigating the effectiveness of 
different training strategies for domain adaptation, our study provides insights for extending MLLMs to other 
vertical domains.

\subsection{Image Captioning in Vertical Domains}
Image captioning, as a representative research task bridging computer vision (CV) and natural language processing 
(NLP), aims to generate relevant and accurate textual descriptions for a given image. This task requires models not 
only to understand objects, scenes, and their relationships within an image\cite{hil} but also to translate this visual 
information into natural and coherent language, thereby jointly evaluating the understanding capabilities of 
computer vision and the generation abilities of natural language processing. Consequently, it has significant 
practical value, and extensive research has emerged across various vertical domains. In the field of news 
reporting, image captioning is used to generate more detailed image descriptions, such as identifying celebrities 
in photographs, thereby enhancing the automation of news content generation and the accuracy of information 
dissemination\cite{ici}. In the biomedical domain, image captioning can assist clinicians in accelerating 
diagnostic processes. Relevant studies summarized datasets, evaluation metrics, and state-of-the-art methods, 
and also proposed baseline models to advance the field\cite{aso}. In underwater photogrammetry, research has 
generated realistic underwater images through physical degradation, and by combining scene–object feature fusion 
models with meta-learning strategies, achieved accurate captioning across diverse underwater scenarios\cite{uic}.

Image captioning in ECE settings also exhibits distinctive characteristics. Compared to general image captioning
tasks, ECE images often involve diverse teaching regions, specialized teaching toys, and interactions between children and teachers, which impose specific requirements on a model’s visual understanding and language generation capabilities. However, datasets and evaluation methods tailored to image captioning in early childhood education remain limited. Therefore, our study focuses on generating captions for daily activity images in ECE settings, aiming to provide systematic research and practical insights for this vertical domain.

\section{ECAC: An Early Childhood Education Daily Activity Captioning Benchmark}

\subsection{Benchmark Dataset Curation}

\subsubsection{Image Collection and Processing}
To address the scarcity of publicly available datasets specifically targeting ECE image captioning, which limits model training and results in inaccurate descriptions of educational toys and daily activities, we introduce the ECAC benchmark dataset. The dataset comprises authentic recordings of children’s daily activities, obtained from routine observations documented by kindergarten teachers. After manually filtering out damaged and blurred images, a total of 256,121 images were retained. Based on the 35 distinct areas established within the kindergarten environment, the images were grouped into three major categories and further classified according to their content. The number of images within each major category is presented in Table. ~\ref{tab:image_distribution}. 

\renewcommand{\arraystretch}{1.5}
\begin{table}
\begin{center}
\caption{Distribution of images across different activity Areas in the dataset.}
\label{tab:image_distribution}
\begin{tabular}{ c  c }
\hline
Category & Number of Images \\
\hline
Outdoor& 56,665 \\
Classroom& 195,397 \\ 
Specialty& 4,059 \\
Total& 256,121 \\
\hline
\end{tabular}
\end{center}
\end{table}

\begin{figure}[h]
    \centering
    \includegraphics[width=0.9\linewidth]{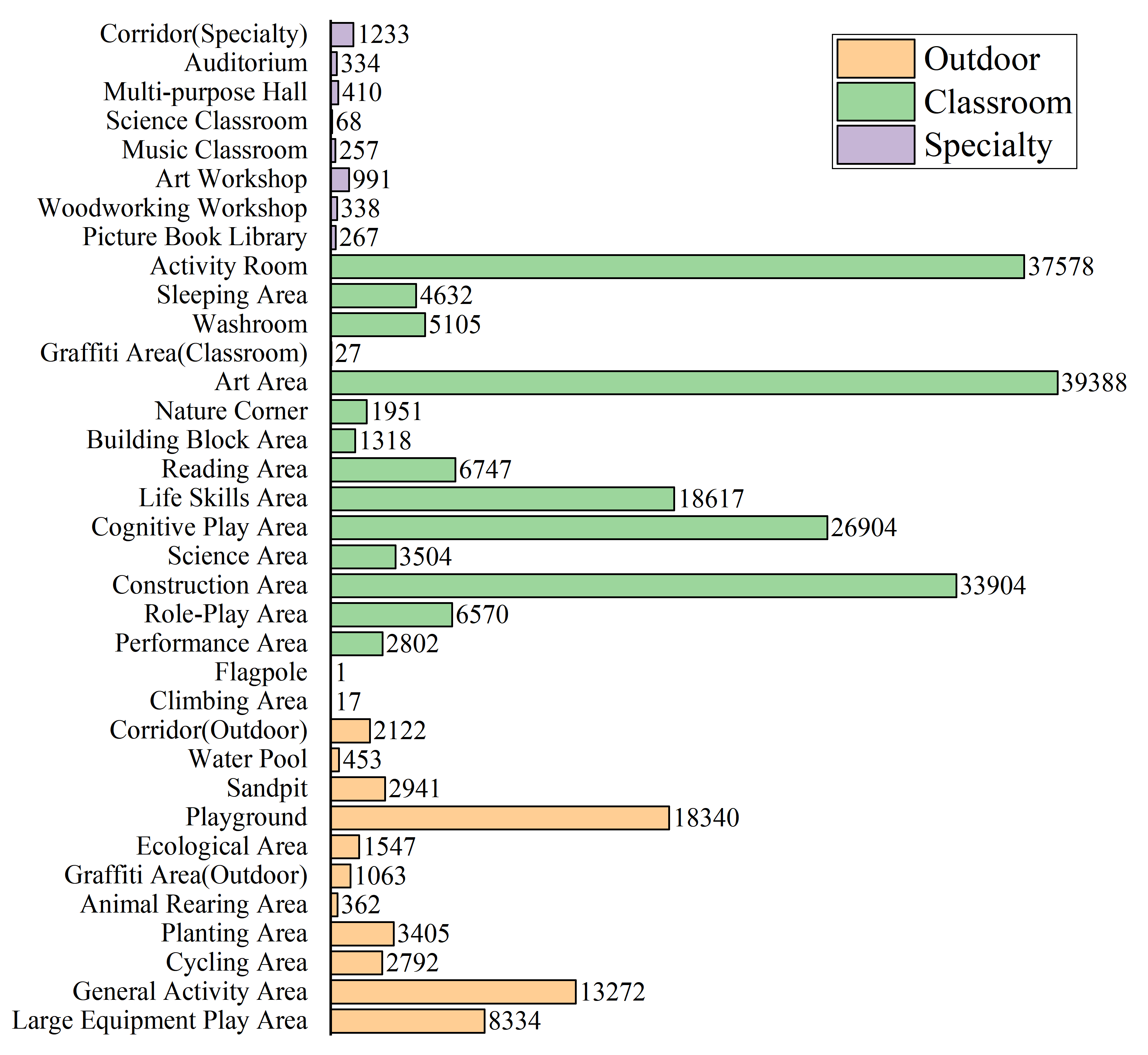} 
    \caption{Distribution of images across different regions in the dataset. }
    \label{fig:distribution_of_images}
\end{figure}

Based on the three major categories, we further subdivided the areas according to their functional differences, ultimately obtaining 35 areas. The name of images for each area are presented in Fig.~\ref{fig:distribution_of_images}.

\subsubsection{Structured Captions Construction Pipeline}
In collaboration with early childhood education experts, we carefully designed a prompt containing three high quality exemplar captions. Based on this prompt, we utilized the GPT-4o API to annotate the images and generate high quality captions.To further enhance the granularity of image annotation, we developed a second prompt with two key functions. First, it identifies the specific names of teaching toys in each image and categorizes their level of domain specificity into low, medium, or high. Second, it classifies teaching toys according to their interaction with individuals, labeling those explicitly involved in interactions as foreground teaching toys and those positioned as background props without noticeable interaction as background teaching toys.To ensure the accuracy and professionalism of the structured annotations, particularly regarding the specific names of teaching toys, we invited eight graduate students specializing in early childhood education research to manually review and refine the model generated annotation structures.This classification strategy ensures that captions accurately reflect the teaching toys actively interacting with individuals, which are more critical than non interactive background props, thereby maintaining a focus on core information and meeting the specific requirements of early childhood education scenarios.

\subsection{Evaluation Metrics Design}

\subsubsection{TTS:Teaching Toy Recognition Score}
The ability to accurately and professionally describe teaching toys appearing in images of children’s daily activities is a critical indicator distinguishing our model from general-purpose models in the context of early childhood education.

To quantitatively evaluate this capability, we adopt a string-matching approach. Specifically, for each image, we take the set of teaching toy names annotated by GPT-4o at different precision levels as the reference, and compare them against the model-generated captions to calculate the match rates for different precision levels. Based on this, we define three basic metrics: TTS-L (match rate for low-precision names), TTS-M (match rate for medium-precision names), and TTS-H (match rate for high-precision names).

\begin{figure*}
    \centering
    \includegraphics[width=\textwidth]{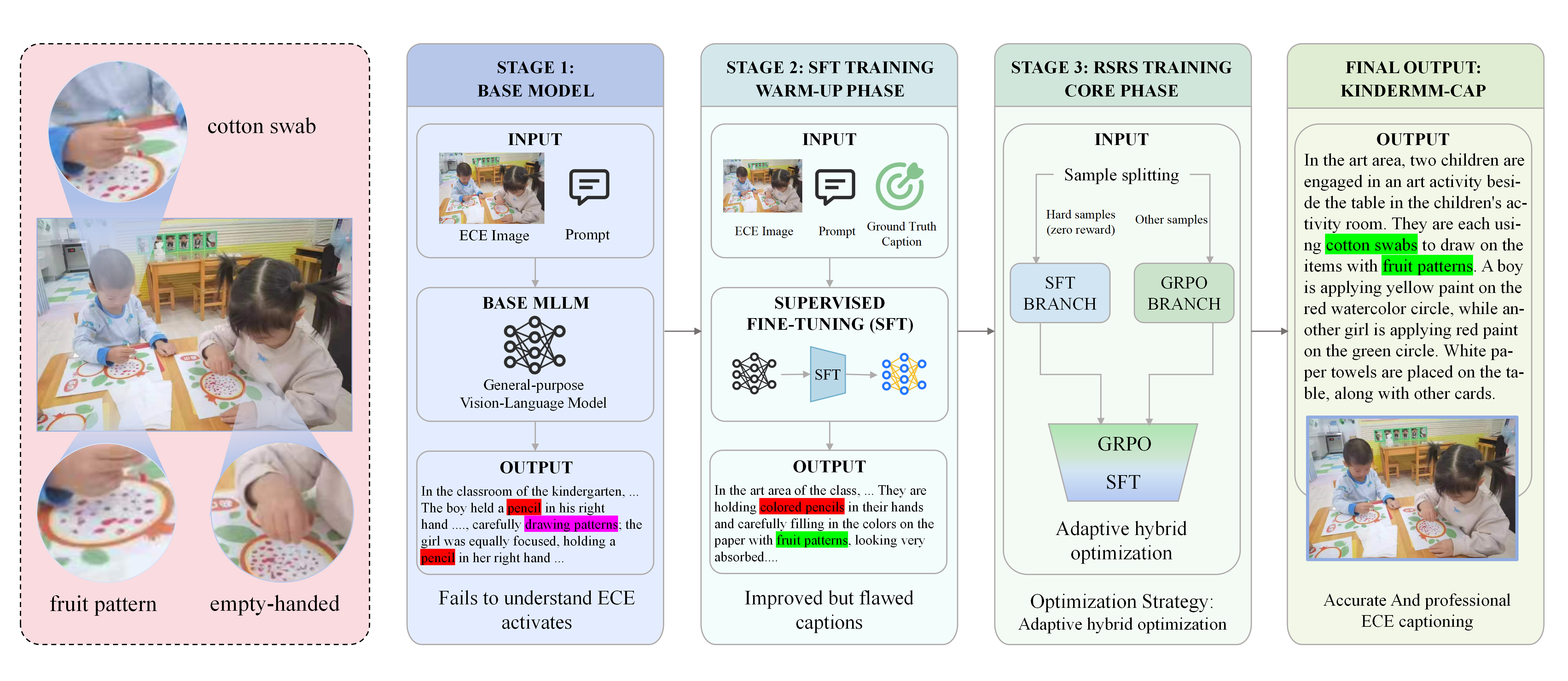} 
    \caption{Overview of the revised KinderMM-Cap framework. The training pipeline comprises three stages: (1) Stage 1 (Base Model), where a general vision–language model generates inaccurate descriptions of Early Childhood Education (ECE) activities; (2) Stage 2 (SFT Warm-up), where supervised fine-tuning improves caption quality but still yields semantic deviations; and (3) Stage 3 (RSRS Core Training), where samples are adaptively split by intra-group rewards—hard samples (zero reward) are optimized via the SFT branch, while others are refined through the GRPO branch to encourage exploration. The final model produces accurate and semantically grounded ECE captions. In the figure, red-highlighted text denotes incorrect descriptions, purple indicates imprecise descriptions, and green represents accurate descriptions.}
    \label{fig:framework-train}
\end{figure*}

Furthermore, leveraging the previously defined “foreground–background” classification of teaching toys, we extend these basic metrics into six fine-grained metrics: TTS-FL (foreground teaching toys with low-precision name matches), TTS-FM (foreground teaching toys with medium-precision name matches), TTS-FH (foreground teaching toys with high-precision name matches), TTS-BL (background  teaching toys with low-precision name matches), TTS-BM (background  teaching toys with medium-precision name matches), and TTS-BH (background teaching toys with high-precision name matches). The computation of each metric can be uniformly expressed as:
\begin{equation}
\delta(a) =
\begin{cases}
1, & \text{match}, \\
0, & \text{no match}.
\end{cases}
\end{equation}
where $\delta(a)$ is an indicator function that equals 1 if the candidate description $a$ correctly matches the toy name, and 0 otherwise.

\begin{equation}
\gamma(t_i) = \mathbf{1}\Big(\sum_{j=1}^{N} \delta(a_{ij}) \ge 1 \Big)
\end{equation}
where $t_i$ is a toy name annotated in the image, $\mathcal{A}_i = \{a_{i1},a_{i2},\dots,a_{iN}\}$ is the set of $N$ candidate descriptions generated for $t_i$, and $\mathbf{1}(\cdot)$ is the indicator function. Thus, $\gamma(t_i)=1$ if at least one candidate description is correct, and 0 otherwise.

\begin{equation}
\mathrm{TTS} = \frac{1}{|T|} \sum_{i=1}^{|T|} \gamma(t_i)
\end{equation}
where $T = \{t_1,t_2,\dots,t_{|T|}\}$ denotes the set of toy names annotated in the image, either in the foreground or background.

\subsubsection{Evaluating Overall Quality of Image Captions with MLLMS}
In addition to performing fine-grained evaluations of scene recognition and educational toy recognition through string-matching methods, we specifically designed prompts to assess the overall quality of image captions from three dimensions: scene recognition, educational toy recognition, and human expression and action recognition. We adopt moonshot-v1-8k-vision-preview as an automated evaluator to assess caption quality. For the detailed prompt specifications, please refer to the following.

Please score the image description according to the content of the picture. The scoring should be based on three aspects:
\begin{itemize}
    \item[1]Whether the names of toys and teaching aids mentioned in the description are accurate and professional—the more accurate and professional, the higher the score.
    \item[2]Whether the description accurately depicts the expressions and actions of the people in the picture.
    \item[3] Whether the description accurately identifies the teaching area where the picture is located, and, if background facilities are mentioned, whether their names are accurate.
\end{itemize}
Please assign a score between 0 and 100. The output must strictly follow the JSON format\{"score":80\}. Do not output anything else.

\section{RSRS: Reward-Conditional Switch of RL and SFT}
We propose a novel two-stage training framework to systematically address the critical limitations of conventional vision-language model training paradigms. These limitations manifest in two aspects: pure supervised learning inherently restricts exploration diversity, while reinforcement learning (RL) often encounters gradient collapse due to vanishing intra-group advantages. The framework first establishes domain knowledge acquisition through supervised learning, followed by the introduction of a Reward-Conditioned Switching Dual-Stream (RSRS) paradigm that synergistically integrates RL with supervised fine-tuning (SFT). Motivated by GRPO methodology, RSRS specifically resolves the persistent challenge of training stagnation through a dynamically adaptive mechanism that switches between dual-stream training modes based on reward signals. This mechanism preserves exploration capacity while mitigating gradient collapse. Subsequently, we formalize the reward function architecture and mathematically characterize the reward-conditioned switching protocol between the dual training streams in the second stage.

\subsection{Overview of the Training Framework}
The framework consists of two training stages: the first is a warm-up phase based on SFT, and the second is the RSRS training stage, which jointly optimizes reinforcement learning (RL) and SFT under a reward-conditioned switching mechanism. 

\begin{figure*}
    \centering
    \includegraphics[width=\textwidth]{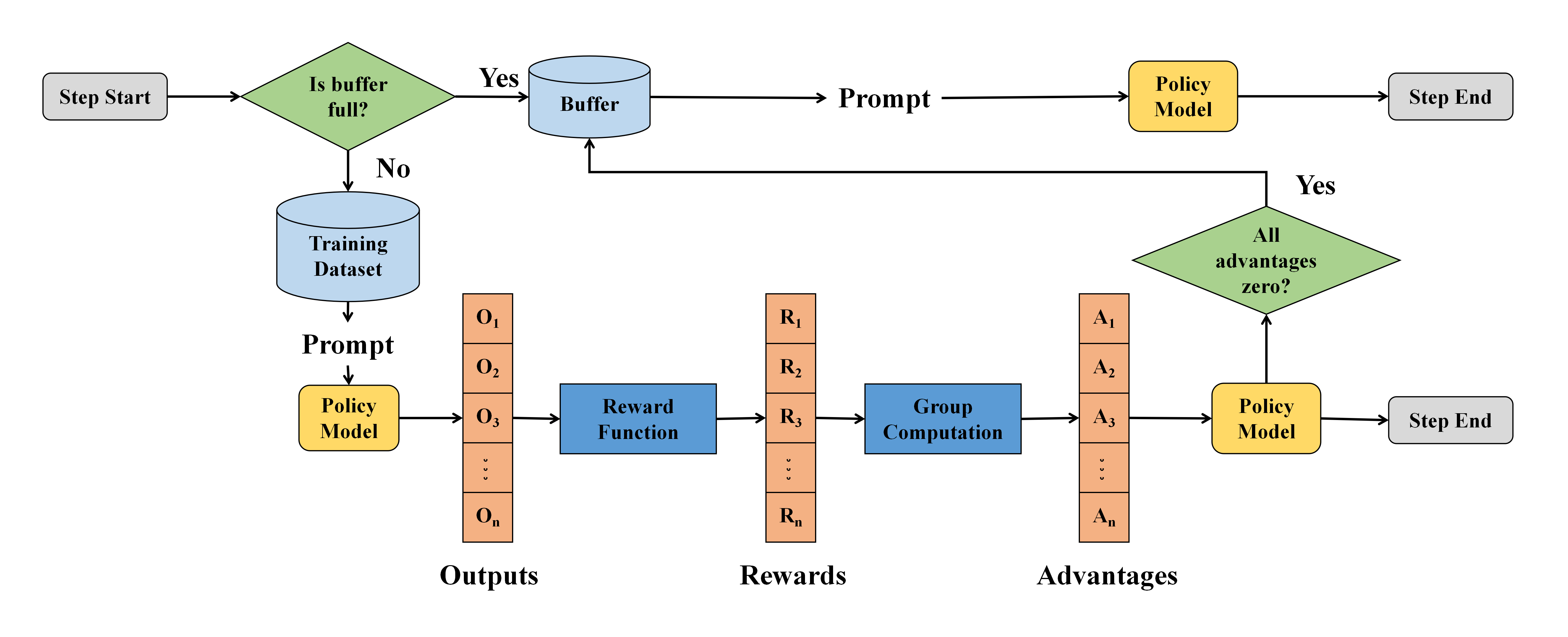} 
    \caption{Diagram of the RSRS Framework. In the diagram, O denotes the candidate captions generated by the model given the input image and prompt; R represents the reward computed from the consistency between the generated captions and annotated teaching toys; A refers to the normalized within-group advantage, which measures the relative quality of each candidate caption compared to the group average. In addition, the buffer stores samples whose entire group receives zero reward, and once it reaches the batch size, the SFT branch is activated to prevent advantage collapse and enhance the model’s learning ability on difficult samples.}
    \label{fig:framework-image}
\end{figure*}

\subsection{Warm-up phase}
The effective generation of high-quality image captions in early childhood education scenarios requires the multimodal large language model (MLLM) to possess robust domain-specific knowledge. However, conventional MLLMs trained on general-domain corpora struggle to capture the nuanced semantics of educational toys, kindergarten activities, and age-appropriate behavioral patterns. To address this limitation, we utilize the ECAC training set, which comprises 247,579 annotated images, to perform domain-adaptive supervised fine-tuning (SFT) on the MLLM. This process enables the model to internalize essential domain characteristics such as pedagogical context, developmental milestones, and activity-specific terminology, thereby significantly enhancing its ability to generate precise and context-aware captions tailored to early childhood education settings.

The core objective of the warm-up stage is to maximize the likelihood of generating authentic image caption sequences by minimizing the negative log-likelihood loss.

\begin{equation}
{{{\mathcal{L}}_{SFT}}}(\theta)=-{{\mathbb{E}}_{(I,p,g)\sim\mathcal{D}}}\left[{\sum_{t=1}^{L}{\log{\pi_\theta}}\left({a_t|s_t}\right)}\right]
\end{equation}

\((I,p,g)\sim D\) denotes sampling from the training data distribution \(D\), where \(I\) represents the input image, \(p\) is the input prompt, and \(g\) is the ground truth caption sequence corresponding to the image, expressed as \(c=\left[a_1,a_2,a_3,...,a_L=<eos>\right]\), with \(a_t\) representing the t-th token in the caption sequence. \(s_t\) denotes the state at step t, with the initial state containing the image features and subsequent states updated with the generated tokens. \(\pi_\theta(a_t|s_t)\) represents the probability of the model generating token \(a_t\) given state \(s_t\). \(L\) is the length of the caption sequence, and \(\theta\)  denotes the model parameters. Through this training paradigm, MLLM not only acquires domain-specific knowledge in early childhood education but also learns to interpret visual elements in images of children’s daily activities and to generate captions with high-quality, structured expressions.

\subsection{RSRS training phase}
Despite effective domain adaptation during warm-up, which primarily focuses on domain-specific knowledge acquisition through supervised fine-tuning (SFT), pure supervised learning remains limited in exploration diversity and fails to adequately address the nuanced professional requirements of educational toy descriptions in early childhood education (ECE) scenarios. While the warm-up SFT stage successfully adapts the model to ECE-specific terminology and context, it struggles with rare, specialized, or visually subtle educational toys that require high-precision recognition. This limitation becomes particularly evident when evaluating the model's performance on challenging samples, where the model often generates inaccurate or generic descriptions of teaching toys (e.g., labeling clay as "craft material" instead of the professional term "modeling clay"). To overcome this constraint, we introduce the RSRS (Reward-Conditional Switch of Reinforcement Learning and Supervised Fine-Tuning) framework, which conducts reward-conditioned adaptive hybrid training by dynamically integrating both SFT and GRPO. In this section, we first introduce the definition of GRPO in the training context, along with the reward functions involved in its training process. Finally, we provide a detailed description of the specific switching mechanism between SFT and GRPO training (see Fig. \ref{fig:framework-image}).

\begin{equation}
\label{equ:grpo}
\begin{aligned}
{{\mathcal{J}}_{GRPO}}(\theta) &= \mathbb{E}\Bigg[
(p,I)\sim D,\; \{o_j\}_{j=1}^G \sim \pi_{\theta_{old}}(O|p,I)
\Bigg] \\
&\frac{1}{G} \sum_{j=1}^{G} \frac{1}{|\theta_j|}
\sum_{t=1}^{|\theta_j|} \min \Bigg[
\frac{\pi_\theta(o_{j,t}|p,I,o_{j<t})}{\pi_{\theta_{old}}(o_{j,t}|p,I,o_{j<t})}\hat{A}_{j,t}, \\
&
\text{clip}\Bigg(
\frac{\pi_\theta(o_{j,t}|p,I,o_{j<t})}{\pi_{\theta_{old}}(o_{j,t}|p,I,o_{j<t})},\; 1\pm \epsilon
\Bigg)\hat{A}_{j,t}
\Bigg]
\end{aligned}
\end{equation}

\subsubsection{Definition of GRPO}
In this stage, GRPO extends policy optimization through inter-group comparisons. For each image \(I\) and prompt input \(p\), \(G\) outputs are sampled from the current policy. Due to computational resource constraints, the reference model is omitted, and thus the KL-divergence constraint is not included in the optimization objective. The optimization objective is defined as Eq. (\ref{equ:grpo}).

\(\pi_{\theta_{old}} \) denotes the previous policy before the update, \(\pi_\theta\) represents the current policy, and \(\frac{\pi_\theta}{\pi_{\theta_{old}}}\) measures the deviation of the new policy from the old policy for each action. \(\hat{A}_{j,t}=\tilde{r_j}=\frac{r_j-\text{mean}(r)}{\text{std}(r)}\) corresponds to the within-group relative advantage, quantifying how much better an action is compared to the average in a given state, \(\epsilon\) controls the clipping range for policy updates.

\subsubsection{Reward function design}
To enhance the MLLM’s proficiency in naming educational toys in captions of daily activity images within the early childhood education domain, we designed a specialized reward function for GRPO training.

\begin{equation}
\label{equ:reward_validation}
    \mathcal{R}(C)=\sum_{i=1}^{n}w_f\cdot{s}(f_i)\cdot {h_{f,i}}+\sum_{j=1}^{m}w_b\cdot {s}(b_i)\cdot {h_{b,j}}
\end{equation}`

In the training set of ECAC, the set \(\Omega\) of teaching toys' names annotated in each image, is divided into the foreground toy set \(\mathcal{F}=\{f_1,f_2,f_3,...,f_n\}\)  and the background toy set \(B=\{b_1,b_2,b_3,...,b_m\}\), such that \(\Omega=F\cup B\) and \(F\cap B={\varnothing} \). Each toy name \(\omega \in \Omega \)  is associated with a precision level function \(\textit{p}(\omega)\in\{\textit{low},\textit{medium},\textit{high}\}\), which is further assigned a corresponding score \(s(p(\omega)\in\{1,2,3\}\). Since the identification of foreground toys is relatively more critical for image captioning, elements in the foreground toy set \(F\) are assigned a weight \(w_f=1\), while elements in the background toy set \(B\) are assigned a weight \(w_b=0.5\). The variable \(h\) denotes whether the teaching toy is present in the caption, taking a value of 1 if present and 0 otherwise. The reward function \(\mathcal{R}(C)\) reflects both the number of toys correctly identified in the image and their level of professionalism.

\begin{algorithm}
\caption{RSRS: Reward-Conditional Switch of RL and SFT}\label{alg:alg1}
\begin{algorithmic}[1]
\REQUIRE
    $D_{\mathrm{SFT}} = \{(I, p, g)\}$: tuples of (image, prompt, ground truth), \\
    $D_{\mathrm{RSRS}} = \{(I, p, g, s)\}$: tuples of (image, question, ground truth, set), \\
    $W$: warm-up steps, \\
    $T$: hybrid training steps, \\
    $\pi_\theta^{(0)}$: initial policy, \\
    $G$: group size,\\
\ENSURE
    ${\pi_\theta}$:final policy
\STATE \text{Initialize }${\pi_\theta=\pi_\theta^{(0)}}$
\STATE \text{// Warm-up stage}
\STATE \textbf{for} $\mathbf{i}=1$ \textbf{to} $W$ \textbf{do}
\STATE \hspace{0.5cm}\text{Sample} $(I,p,g)\sim D_{SFT}$
\STATE \hspace{0.5cm}\text{Update} $\theta = \text{OPTIMIZATION\_STEP}(L_{SFT}(\theta))$
\STATE \textbf{end for}
\STATE \text{// Hybrid training stage}
\STATE \textbf{for} $\mathbf{t}=1$ \textbf{to} $T$ \textbf{do}
\STATE \hspace{0.5cm}\textbf{if} $\text{Buffer Size}<\text{Batch Size}$ \textbf{then}
\STATE \hspace{1cm}\text{Sample} $(I,p,g,s)\sim D_{RSRS}$
\STATE \hspace{1cm}\text{Generate} $\{o_i\}_{i=1}^G \gets \pi_{\theta}(I,p)$
\STATE \hspace{1cm}\text{Compute rewards} $\{r_i\}_{i=1}^G \gets \mathcal{R}(\{o_i\}_{i=1}^G)$
\STATE \hspace{1cm}\textbf{if} $\text{sum}(\{r_i\}_{i=1}^G)=0$ \textbf{then}
\STATE \hspace{1.5cm}\text{Buffer Append} $(I,p,g,s)$
\STATE \hspace{1cm}\textbf{end if}
\STATE \hspace{1cm}\text{Update} ${\theta = \text{OPTIMIZATION\_STEP}(L_{GRPO}(\theta))}$
\STATE \hspace{0.5cm}\textbf{else}
\STATE \hspace{1cm}\text{Sample} $(I,p,g)\sim $ \text{Buffer}
\STATE \hspace{1cm}\text{Buffer Pop} $(I,p,g,s)$
\STATE \hspace{1cm}\text{Update} $\theta = \text{OPTIMIZATION\_STEP}(L_{SFT}(\theta))$
\STATE \hspace{0.5cm}\textbf{end if}
\STATE \textbf{end for}
\STATE \textbf{return} $\pi_\theta$
\end{algorithmic}
\label{alg1}
\end{algorithm}

\subsubsection{Reward condition-based switching mechanism}
Our training framework innovatively integrates SFT with GRPO through a reward-based switching mechanism. The core contribution lies in addressing a critical challenge in GRPO training: when encountering difficult samples, all G responses generated by the model either receive no reward or are assigned identical rewards. This leads to advantage values collapsing to zero, gradient vanishing, and ultimately results in training stagnation and restricted exploration. 

The absence of rewards indicates that the model’s current capabilities are insufficient to meet the demands posed by certain challenging samples—an observation that underscores the intrinsic value of such samples for improving model performance. To address this issue, we propose a mechanism, illustrated in Fig. \ref{fig:framework-image}: when all responses within a group receive zero reward, the corresponding data sample is stored in a buffer. Once the number of buffered samples reaches or exceeds the training batch size, a batch is sampled from the buffer, and the training mode automatically switches from GRPO to SFT, thereby effectively harnessing the learning potential of these difficult samples. Based on the above mechanism, the overall procedure of the RSRS algorithm can be summarized as follows:


\begin{thebibliography}{1}
\bibliographystyle{IEEEtran}

\bibitem{rcmf}
L. Wang, H. Chen, Y. Liu and Y. Lyu, "Regular Constrained Multimodal Fusion for Image Captioning," in IEEE Transactions on Circuits and Systems for Video Technology, vol. 34, no. 11, pp. 11900-11913, Nov. 2024. 

\bibitem{fla}
Alayrac J B, Donahue J, Luc P, et al. Flamingo: a visual language model for few-shot learning[J]. Advances in neural information processing systems, 2022, 35: 23716-23736.

\bibitem{mini}
Zhu D, Chen J, Shen X, et al. Minigpt-4: Enhancing vision-language understanding with advanced large language models[J]. arXiv preprint arXiv:2304.10592, 2023.

\bibitem{llm}
Brown T, Mann B, Ryder N, et al. Language models are few-shot learners[J]. Advances in neural information processing systems, 2020, 33: 1877-1901.

\bibitem{tlm}
Ouyang L, Wu J, Jiang X, et al. Training language models to follow instructions with human feedback[J]. Advances in neural information processing systems, 2022, 35: 27730-27744.

\bibitem{hfb}
Jin Q, Chen F, Zhou Y, et al.Hidden flaws behind expert-level accuracy of multimodal GPT-4 vision in medicine[J].NPJ Digital Medicine, 2024, 7(1): 190.

\bibitem{pml}
Zhou J, He X, Sun L, et al.Pre-trained multimodal large language model enhances dermatological diagnosis using SkinGPT-4[J].Nature Communications, 2024, 15(1): 5649.

\bibitem{fin}
Gagan Bhatia, El Moatez Billah Nagoudi, Hasan Cavusoglu, and Muhammad Abdul-Mageed.2024.FinTral: A Family of GPT-4 Level Multimodal Financial Large Language Models.In Findings of the Association for Computational Linguistics: ACL 2024, pages 13064–13087, Bangkok, Thailand.Association for Computational Linguistics.

\bibitem{mme}
Gan Z, Lu Y, Zhang D, et al. Mme-finance: A multimodal finance benchmark for expert-level understanding and reasoning[J]. arXiv preprint arXiv:2411.03314, 2024.

\bibitem{cfb}
Li J, Zhu Y, Cheng D, et al. CFBenchmark-MM: Chinese Financial Assistant Benchmark for Multimodal Large Language Model[J]. arXiv preprint arXiv:2506.13055, 2025.

\bibitem{dfp}
Y. Tang, L. Zhang, Y. Yuan and Z. Chen, "Describe Fashion Products via Local Sparse Self-Attention Mechanism and Attribute-Based Re-Sampling Strategy," in IEEE Transactions on Circuits and Systems for Video Technology, vol. 33, no. 7, pp. 3409-3424, July 2023.

\bibitem{agr}
Awais M, Alharthi A H S A, Kumar A, et al. Agrogpt: Efficient agricultural vision-language model with expert tuning[C]//2025 IEEE/CVF Winter Conference on Applications of Computer Vision (WACV). IEEE, 2025: 5687-5696.

\bibitem{isy}
Lee U, Jeong Y, Koh J, et al. I see you: teacher analytics with GPT-4 vision-powered observational assessment[J]. Smart Learning Environments, 2024, 11(1): 48.

\bibitem{etp}
Xu S, Huang X, Lo C K, et al. Evaluating the performance of ChatGPT and GPT-4o in coding classroom discourse data: A study of synchronous online mathematics instruction[J]. Computers and Education: Artificial Intelligence, 2024, 7: 100325.

\bibitem{etpo}
Yu S, Androsov A, Yan H. Exploring the prospects of multimodal large language models for Automated Emotion Recognition in education: Insights from Gemini[J]. Computers \& Education, 2025, 232: 105307.

\bibitem{ner}
Lee G G, Zhai X. Nerif: Gpt-4v for automatic scoring of drawn models[J]. arXiv preprint arXiv:2311.12990, 2023.

\bibitem{gpd}
Lee G G, Latif E, Shi L, et al. Gemini pro defeated by gpt-4v: Evidence from education[J]. arXiv preprint arXiv:2401.08660, 2023.

\bibitem{doc}
Lee U, Jeon M, Lee Y, et al.LLaVA-docent: Instruction tuning with multimodal large language model to support art appreciation education[J].Computers and Education: Artificial Intelligence, 2024, 7: 100297.

\bibitem{tml}
Shi W, Le H V, Choo K T W. Towards multimodal large-language models for parent-child interaction: A focus on joint attention[C]//Proceedings of the Extended Abstracts of the CHI Conference on Human Factors in Computing Systems. 2025: 1-6.

\bibitem{dino}
Wang B, Song Y, Cao J, et al. DinoCompanion: An Attachment-Theory Informed Multimodal Robot for Emotionally Responsive Child-AI Interaction[J]. arXiv preprint arXiv:2506.12486, 2025.

\bibitem{deep}
Shao Z, Wang P, Zhu Q, et al.Deepseekmath: Pushing the limits of mathematical reasoning in open language models[J].arXiv preprint arXiv:2402.03300, 2024.

\bibitem{aat}
Xu K, Ba J, Kiros R, et al. Show, attend and tell: Neural image caption generation with visual attention[C]//International conference on machine learning. PMLR, 2015: 2048-2057.

\bibitem{mde}
Chen X, Lawrence Zitnick C. Mind's eye: A recurrent visual representation for image caption generation[C]//Proceedings of the IEEE conference on computer vision and pattern recognition. 2015: 2422-2431.

\bibitem{hai}
Dong G, Yuan H, Lu K, et al. How abilities in large language models are affected by supervised fine-tuning data composition[J]. arXiv preprint arXiv:2310.05492, 2023.

\bibitem{spt}
H. Zhang, P. Zeng, L. Gao, X. Lyu, J. Song and H. T. Shen, "SPT: Spatial Pyramid Transformer for Image Captioning," in IEEE Transactions on Circuits and Systems for Video Technology, vol. 34, no. 6, pp. 4829-4842, June 2024.

\bibitem{blip}
Liu H, Li C, Li Y, et al. Improved baselines with visual instruction tuning[C]//Proceedings of the IEEE/CVF conference on computer vision and pattern recognition. 2024: 26296-26306.

\bibitem{qw25}
Bai S, Chen K, Liu X, et al. Qwen2. 5-vl technical report[J]. arXiv preprint arXiv:2502.13923, 2025.

\bibitem{VQA}
Goyal Y, Khot T, Summers-Stay D, et al. Making the v in vqa matter: Elevating the role of image understanding in visual question answering[C]//Proceedings of the IEEE conference on computer vision and pattern recognition. 2017: 6904-6913.

\bibitem{okvqa}
Marino K, Rastegari M, Farhadi A, et al. Ok-vqa: A visual question answering benchmark requiring external knowledge[C]//Proceedings of the IEEE/cvf conference on computer vision and pattern recognition. 2019: 3195-3204.

\bibitem{docvqa}
Mathew M, Karatzas D, Jawahar C V. Docvqa: A dataset for vqa on document images[C]//Proceedings of the IEEE/CVF winter conference on applications of computer vision. 2021: 2200-2209.

\bibitem{ocrb}
Liu Y, Li Z, Huang M, et al. Ocrbench: on the hidden mystery of ocr in large multimodal models[J]. Science China Information Sciences, 2024, 67(12): 220102.

\bibitem{msr}
Xu J, Mei T, Yao T, et al. Msr-vtt: A large video description dataset for bridging video and language[C]//Proceedings of the IEEE conference on computer vision and pattern recognition. 2016: 5288-5296.

\bibitem{oven}
Hu H, Luan Y, Chen Y, et al. Open-domain visual entity recognition: Towards recognizing millions of wikipedia entities[C]//Proceedings of the IEEE/CVF International Conference on Computer Vision. 2023: 12065-12075.

\bibitem{vrs}
Li X, Ding J, Elhoseiny M. Vrsbench: A versatile vision-language benchmark dataset for remote sensing image understanding[J]. Advances in Neural Information Processing Systems, 2024, 37: 3229-3242.

\bibitem{med}
Zong Y, Yang Y, Hospedales T. MEDFAIR: benchmarking fairness for medical imaging[J]. arXiv preprint arXiv:2210.01725, 2022.

\bibitem{coco}
Lin T Y, Maire M, Belongie S, et al. Microsoft coco: Common objects in context[C]//European conference on computer vision. Cham: Springer International Publishing, 2014: 740-755.

\bibitem{vis}
Liu Z, Guo X, Xia H, et al. VisFinEval: A Scenario-Driven Chinese Multimodal Benchmark for Holistic Financial Understanding[J]. arXiv preprint arXiv:2508.09641, 2025.

\bibitem{amo}
Ji Y, Bai H, Ge C, et al. Amos: A large-scale abdominal multi-organ benchmark for versatile medical image segmentation[J]. Advances in neural information processing systems, 2022, 35: 36722-36732.

\bibitem{tka}
Yi H, Liu T, Lan G. The key artificial intelligence technologies in early childhood education: a review[J]. Artificial Intelligence Review, 2024, 57(1): 12.

\bibitem{acv}
Del Coco M, Leo M, Carcagni P, et al. A computer vision based approach for understanding emotional involvements in children with autism spectrum disorders[C]//Proceedings of the IEEE international conference on computer vision workshops. 2017: 1401-1407.

\bibitem{cul}
Rudovic O, Utsumi Y, Lee J, et al. CultureNet: a deep learning approach for engagement intensity estimation from face images of children with autism[C]//2018 IEEE/RSJ International Conference on Intelligent Robots and Systems (IROS). IEEE, 2018: 339-346.

\bibitem{irf}
Dongming L, Wanjing L, Shuang C, et al. Intelligent robot for early childhood education[C]//Proceedings of the 2020 8th international conference on information and education technology. 2020: 142-146.

\bibitem{dso}
Xia Y, Huang D, Wang Y. Detecting smiles of young children via deep transfer learning[C]//Proceedings of the IEEE international conference on computer vision workshops. 2017: 1673-1681.

\bibitem{fli}
Plummer B A, Wang L, Cervantes C M, et al. Flickr30k entities: Collecting region-to-phrase correspondences for richer image-to-sentence models[C]//Proceedings of the IEEE international conference on computer vision. 2015: 2641-2649.

\bibitem{hil}
Y. Wang, N. Xu, A. -A. Liu, W. Li and Y. Zhang, "High-Order Interaction Learning for Image Captioning," in IEEE Transactions on Circuits and Systems for Video Technology, vol. 32, no. 7, pp. 4417-4430, July 2022

\bibitem{ici}
Liu T, Cai Q, Xu C, et al.Image Captioning in news report scenario[J].arXiv preprint arXiv:2403.16209, 2024.

\bibitem{aso}
Pavlopoulos J, Kougia V, Androutsopoulos I.A survey on biomedical image captioning[C]//Proceedings of the second workshop on shortcomings in vision and language.2019: 26-36.

\bibitem{uic}
Li H, Wang H, Zhang Y, et al.Underwater image captioning: Challenges, models, and datasets[J].ISPRS Journal of Photogrammetry and Remote Sensing, 2025, 220: 440-453.

\end{thebibliography}
%

\section{EXPERIMENTS}
\subsection{Experimental Setups}
\subsubsection{Models and Datasets}Due to the strong performance of the Qwen2.5-VL series in Chinese tasks, we selected Qwen2.5-VL-Instruction-3B as the backbone model. All subsequent training in this work was conducted based on this model. In addition, we evaluated mainstream open-source MLLMs. In the warm-up stage, the training data were derived from 247,579 image-text pairs in the ECAC dataset. In the hybrid training phase, the mixed training data were composed of 12,000 images from ECAC, together with annotated sets of teaching toy names with varying levels of precision and the corresponding ground-truth captions.
\renewcommand{\arraystretch}{1.5}
\begin{table}[h]
\begin{center}
\caption{Performance Comparison Table of MLLMs for Daily Activity Image Captioning in ECE}
\label{tab:main result}
\begin{tabular}{ c  c  c }
\hline
Model & TTS & Score \\
\hline
GLM-4v-9B& 29.36& 78.19 \\
Gemma3-27B& 10.74 & 12 \\ 
Qwen2.5-VL-3B& 38.07 & 84.63 \\
Qwen2.5-VL-7B&45.25 &85.70 \\
KinderMM-Cap-3B(ours)& \textbf{51.06}& 86.20 \\
\hline
\end{tabular}
\end{center}
\end{table}
\subsubsection{Setups}We adopted LLaVA Factory to conduct SFT on Qwen2.5-VL-Instruction-3B during the warm-up stage. The training was carried out on 8 NVIDIA V100 GPUs for a total of 3 epochs, with the learning rate set to 1.0e-5. This stage was designed to stabilize the model parameters and adapt the backbone to the downstream instruction-following task.
In the hybrid training phase, the same hardware configuration of 8 V100 GPUs was employed; however, the computation was split across two groups. Specifically, 4 GPUs were allocated for training, while the remaining 4 GPUs were used for online inference through vLLM. In this setup, the inference side generated 4 candidate responses per step, and the training side updated the model accordingly, which resulted in an effective global batch size of 4. This design allowed the model to iteratively benefit from both training signals and online-generated supervision.

\subsection{Reward Function Validation}

To verify that the proposed reward function aligns with human 
expert judgment in the ECE domain, we conducted a human 
preference correlation study. Specifically, for each of 100 
images randomly sampled from the ECAC test set, five captions 
were collected from the following models: Qwen2.5-VL-3B-Instruct, 
Qwen2.5-VL-7B-Instruct, Qwen2.5-VL-7B-Instruct (SFT), 
Qwen2.5-VL-32B-Instruct, and Qwen2.5-VL-72B-Instruct. 
These models were deliberately selected from the Qwen2.5-VL 
family across varying parameter scales to ensure sufficient 
quality diversity among the five captions, thereby facilitating 
more discriminative and reliable human ranking.

Four graduate students with expertise in early childhood 
education were recruited as human annotators. Each annotator 
independently ranked the five captions per image based solely 
on the accuracy and professionalism of educational toy naming, 
without prior knowledge of the source model. The averaged 
rankings across annotators were taken as the ground-truth 
preference signal.

We evaluated four automatic ranking methods against these 
human rankings: (1)~\textbf{CN-CLIP}, which orders captions 
by image-text similarity score; (2)~\textbf{GPT-4o}, 
a proprietary vision-language model employed as a scoring 
judge; (3)~\textbf{Gemma3-27B}, an open-source model employed 
as a scoring judge; and (4)~\textbf{Ours}, the proposed reward 
function $\mathcal{R}(C)$ (Eq.~\ref{equ:reward_validation}), which scores 
captions based on annotated teaching toy name matching across 
multiple precision levels.

Ranking consistency with human judgments is quantified using 
Kendall's $\tau$ and Spearman's $\rho$ correlation coefficients. 
As reported in Table. \ref{tab:reward_validation}, the proposed 
reward function achieves the highest agreement with human 
rankings ($\tau = 0.67$, $\rho = 0.72$), substantially 
outperforming CN-CLIP ($\tau = 0.11$, $\rho = 0.13$) and 
both model-based judges. These results demonstrate that 
$\mathcal{R}(C)$ effectively captures domain-specific caption 
quality criteria that are consistent with expert human 
perception in ECE settings, thereby validating its suitability 
as a reward signal for reinforcement learning in our framework.

\renewcommand{\arraystretch}{1.5}
\begin{table}[h]
\begin{center}
\caption{Correlation Between Automatic Ranking Methods 
         and Human Judgments.}
\label{tab:reward_validation}
\begin{tabular}{ c  c  c }
\hline
\textbf{Method} & \textbf{Kendall's $\tau$} 
                & \textbf{Spearman's $\rho$} \\
\hline
CN-CLIP                          & 0.11 & 0.13 \\
GPT-4o   & 0.43 & 0.52 \\
Gemma3-27B                       & 0.26 & 0.32 \\
Ours                             & \textbf{0.67} & \textbf{0.72} \\
\hline
\end{tabular}
\end{center}
\end{table}

\subsection{Main Results}
This experiment adopts Qwen2.5-VL-Instruction-3B as the backbone model to construct KinderMM-Cap-3B. Based on the self-developed ECAC dataset (256,121 ECE daily activity images) and the RSRS dual-track training framework (dynamically integrating SFT and GRPO), it focuses on two core metrics: Teaching Toy Recognition Score (TTS) and overall caption quality, comparing with mainstream MLLMs such as GLM-4v-9B and Gemma3-27B.

As shown in Table. \ref{tab:main result}, KinderMM-Cap-3B achieves a TTS of 51.06, significantly outperforming all comparative models—34.1\% higher than the baseline Qwen2.5-VL-3B (38.07), and superior to the larger-scale Qwen2.5-VL-7B (45.25) and 32B (29.96). General-purpose models perform poorly (Gemma3-27B only 10.74), verifying the necessity of domain-specific data and targeted training frameworks.

KinderMM-Cap-3B scores 86.20, slightly lower than Qwen2.5-VL-32B (89.13) but achieving a balance between core performance and comprehensive quality at the 3B parameter scale, far exceeding GLM-4v-9B (78.19) and Gemma3-27B (12.00).

The fine-grained annotations of the ECAC dataset and the RSRS framework effectively address the insufficient accuracy of teaching toy recognition in ECE scenarios; 2. Targeted optimization frameworks are more efficient than pure parameter expansion, providing a low-cost solution for resource-constrained scenarios; 3. The proposed model can generate professional and high-quality image captions, offering technical support for the intelligent transformation of ECE.

\subsection{Ablation Study}
\begin{figure}[h]
    \centering
    \includegraphics[width=0.9\linewidth]{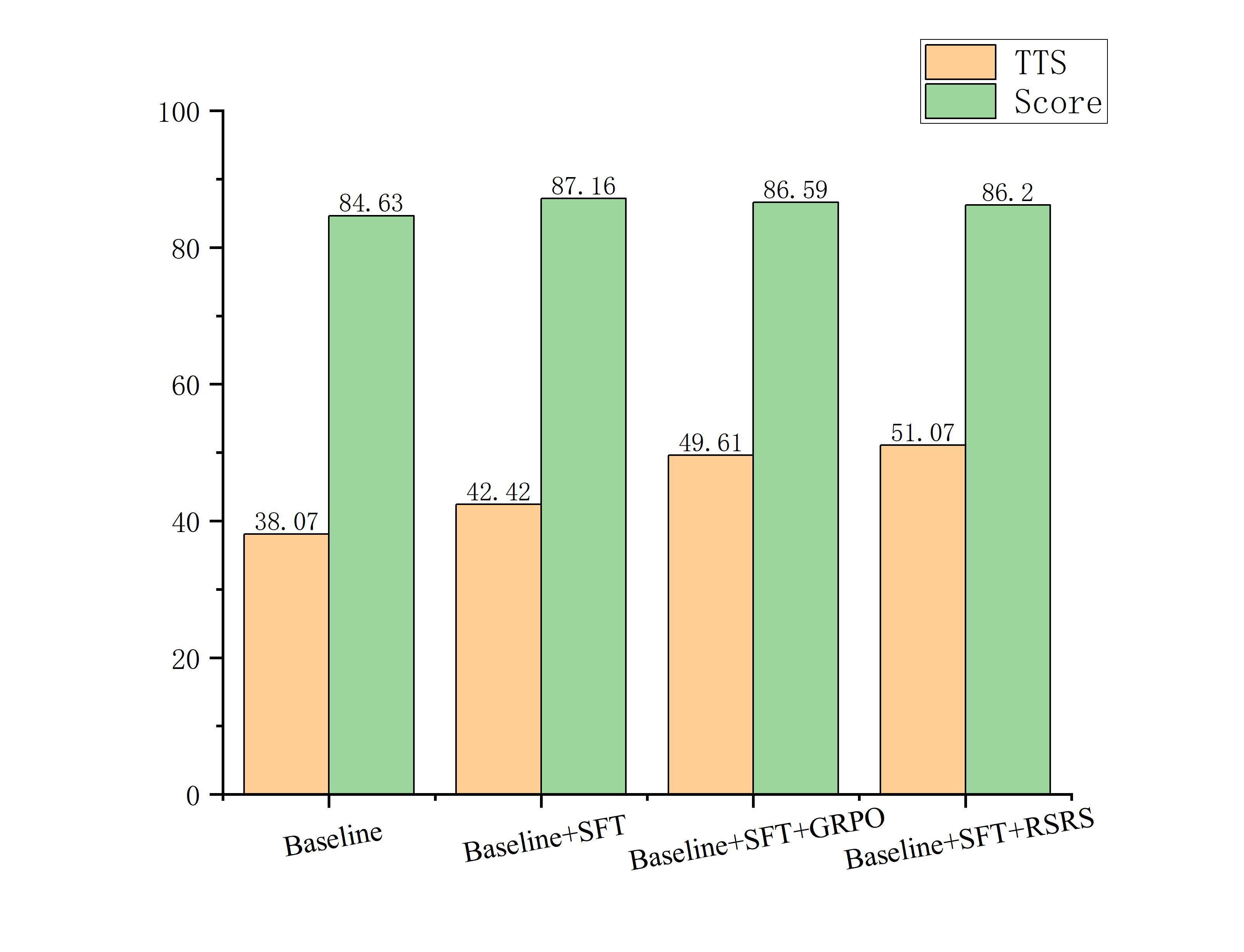} 
    \caption{Ablation Study on Training Strategies for ECE Daily Activity Captioning. }
    \label{fig:ablation study toy}
\end{figure}

As shown in Fig. \ref{fig:ablation study toy}, SFT with 247,579 samples improved TTS by 11.3\% (42.42) over the baseline. However, GRPO with only 12,000 samples (4.8\% of SFT data) achieved a 17.0\% improvement over SFT (49.61 TTS), demonstrating its reward-driven learning mechanism's superior data efficiency. The proposed RSRS framework further enhanced TTS to 51.07 (2.9\% relative improvement over GRPO) while maintaining competitive caption quality (Score=86.20 vs. 86.59).

\renewcommand{\arraystretch}{1.5}
\begin{table*}[t]
\begin{center}
\caption{Multi-Dimensional Evaluation of Toy Recognition Metrics Across Training Methods}
\label{tab:ablation study detail}
\begin{tabular}{ c  c  c  c  c  c  c }
\hline
Model & TTS-FL & TTS-FM &TTS-FH &TTS-BL &TTS-BM &TTS-BH  \\
\hline
baseline& 57.47 & 8.88 & 0.03 & 24.52 & 1.53 & 0.01\\
baseline+SFT& \textbf{68.21} & 15.71  & 0.04 & 24.15 & 2.01 & 0.0\\ 
baseline+SFT+GRPO& 65.66 & 14.39  & 0.02 & 37.97 & 2.86 & 0.02\\
baseline+SFT+RSRS& 66.59 & \textbf{17.18}  & \textbf{0.09} & \textbf{39.69} & \textbf{3.87} & \textbf{0.03}\\
\hline
\end{tabular}
\end{center}
\end{table*}

As shown in Table. \ref{tab:ablation study detail}, RSRS surpasses pure GRPO across nearly all toy-recognition metrics, with particularly large gains in medium-/high-precision and background toy categories. GRPO performs reasonably only on low-precision foreground toys (TTS-FL = 65.66) but exhibits almost no capacity to recognize high-precision toys, with TTS-FH and TTS-BH both remaining at 0.02. Background toy recognition also shows clear limitations under GRPO (TTS-BL = 37.97, TTS-BM = 2.86). These results reveal a structural issue in GRPO’s reward-driven optimization: when all G sampled outputs fail to identify any toy, the group receives zero reward, causing advantage collapse and preventing gradient updates. Consequently, GRPO tends to learn only easy, frequent toy names while failing to improve on rare, specialized, or visually subtle toy categories.

In contrast, RSRS demonstrates notable improvements across all precision levels, especially on components where GRPO struggled. High-precision recognition improves from 0.02 to 0.09 on TTS-FH and from 0.02 to 0.03 on TTS-BH. Medium-precision indicators also increase significantly (TTS-FM: 14.39 → 17.18; TTS-BM: 2.86 → 3.87). These gains stem directly from RSRS’s reward-conditioned dual-stream mechanism: GRPO is applied to samples with non-zero rewards, while groups with zero reward are routed into a buffer and subsequently trained with supervised fine-tuning (SFT). Since high-precision and background toys frequently yield zero-reward groups, they trigger SFT more often, providing ground-truth gradients that GRPO alone cannot supply. This resolves the zero-gradient zone inherent in GRPO and substantially enhances recognition of domain-specific and long-tail toy categories.

Overall, RSRS combines reinforcement learning–driven exploration with supervised correction, producing balanced improvements across foreground/background and low/medium/high precision levels. For instance, TTS-BL increases from 37.97 under GRPO to 39.69 under RSRS, while TTS-FL improves slightly to 66.59, demonstrating consistent enhancement rather than mode collapse toward easy categories. This dual optimization pathway fundamentally reshapes the learning dynamics, enabling RSRS to outperform GRPO across all key indicators reported in Table. \ref{tab:ablation study detail}.

To verify whether models insert a large number of irrelevant toys into descriptions to gain reward signals, we conducted an experiment evaluating toy recognition precision. The results in Table. \ref{tab:ablation study pre} reveal critical insights into this behavior. In the context of toy recognition in ECE captioning, precision quantifies the accuracy of predicted toy names relative to ground truth annotations. It is formally defined as the ratio of correctly identified toys (true positives) to the total number of predicted toys, including both correct and incorrect identifications. The mathematical formulation is:

\begin{equation}
    \text{Precision} = \frac{|\mathcal{T}_{\text{caption}} \cap \mathcal{T}_{\text{groundtruth}}|}{|\mathcal{T}_{\text{caption}}|}
\label{eq:precision}
\end{equation}

Where $T_{caption}$ is the set of toy names predicted by the model in the caption. These predictions are extracted using a large language model(LLM), specifically Qwen2.5-7B-Instruct, which identifies and parses toy names from generated captions $T_{groundtruth}$ is the set of actual toy names present in the image. $|\cdot|$ denotes the cardinality (number of elements) of a set.

The baseline model (1.45 predicted toys, 64.15\% precision) demonstrates limited capability in both quantity estimation and object recognition. SFT improves the predicted toy count to 1.66 while achieving a 69.59\% precision rate, indicating its effectiveness in domain-specific knowledge acquisition. However, the RSRS framework, which dynamically switches between reinforcement learning and supervised learning, increases the predicted toy count to 2.08 at the cost of a significant precision drop to 56.43\%. This suggests that reward-driven optimization may encourage models to insert irrelevant toys to maximize reward signals. In contrast, the GRPO method achieves a balanced performance with 1.84 predicted toys and 64.43\% precision, maintaining stability in both metrics.

The RSRS framework increases the average predicted toy count by 43.4\%, accompanied by a 14.1\% decrease in precision. This suggests that reward-driven optimization may encourage more aggressive toy prediction behavior. The 14.4\% improvement in predicted toy count and 5.4\% precision gain confirm SFT's effectiveness in adapting general-purpose models to ECE-specific tasks without encouraging over-prediction. GRPO's performance (1.84/64.43\%) shows only marginal improvements over baseline+SFT (1.66/69.59\%), but maintains stability in precision, suggesting its potential as a regularization strategy against reward-driven overfitting.

\renewcommand{\arraystretch}{1.5}
\begin{table}[h]
\begin{center}
\caption{Toy Recognition Precision in ECE Captioning with Different Training Strategies}
\label{tab:ablation study pre}
\begin{tabular}{ c  c  c }
\hline
Model & Avg. Toy Count & Precision \\
\hline
baseline& 1.45 & 64.15\\
baseline+SFT& 1.66  & 69.59\\ 
baseline+SFT+GRPO& 1.84 & 64.43\\
baseline+SFT+RSRS& 2.08 &56.43 \\
\hline
\end{tabular}
\end{center}
\end{table}

These findings highlight the need for hybrid approaches that can simultaneously optimize both quantity estimation and naming accuracy while preventing reward-induced over-prediction. The baseline+SFT combination currently achieves the best balance, while RSRS's tendency to prioritize reward signals over factual accuracy requires further investigation.

\section{CONCLUSION}
This paper proposes a comprehensive solution for the image captioning task of daily activity images in early childhood education (ECE). We construct the first ECE-specific benchmark dataset, ECAC, and design the RSRS training framework to effectively address the issue of inaccurate educational toy recognition. Experimental results demonstrate that the RSRS framework, by dynamically integrating reinforcement learning and supervised fine-tuning, significantly improves the professional recognition rate of educational toys while maintaining high-quality descriptions. The trained model, KinderMM-Cap-3B, achieves a TTS score of 51.06, 34.1\% higher than the baseline models. Furthermore, our study reveals the relationship between data efficiency and algorithmic effectiveness, proving that a combination of small-scale high-quality datasets and targeted training frameworks can achieve comparable performance to large-scale data training. Future work will explore more refined description generation strategies and extend this method to more ECE scenarios, such as emotion recognition and social interaction analysis of young children, providing comprehensive technical support for intelligent education.

\vfill  %
\end{CJK*}
\end{document}